\pdfoutput=1

\documentclass[11pt]{article}
\usepackage[utf8]{inputenc}
\usepackage[T1]{fontenc}

\usepackage[]{ACL2023}

\usepackage{times}
\usepackage{latexsym}

\usepackage[T1]{fontenc}

\usepackage[utf8]{inputenc}

\usepackage{microtype}

\usepackage{inconsolata}

\usepackage{booktabs}
\usepackage{graphicx}
\usepackage{amsmath}
\usepackage{subcaption}

\title{Improving Low-Resource Morphological Inflection via Self-Supervised Objectives}

\author{
    Adam Wiemerslage$^{1,3}$ \and Katharina von der Wense$^{1,2}$ \\
    $^1$University of Colorado Boulder \\
    $^2$Johannes Gutenberg University Mainz \\
    $^3$Kensho Technologies \\
    \texttt{\{adwi9965,katharina.kann\}@colorado.edu}
}
\begin{document}
\maketitle
\begin{abstract}
Self-supervised objectives have driven major advances in NLP by leveraging large-scale unlabeled data, but such resources are scarce for many of the world’s languages. Surprisingly, they have not been explored much for
character-level tasks, 
where smaller amounts of data have the potential to be beneficial.
We investigate the effectiveness of self-supervised auxiliary tasks for morphological inflection -- a character-level task highly relevant for language documentation -- in extremely low-resource settings, training encoder-decoder transformers for 19 languages and 13 auxiliary objectives. Autoencoding yields the best performance when unlabeled data is very limited, while character masked language modeling (CMLM) becomes more effective as data availability increases.
Though objectives with stronger inductive biases influence model predictions intuitively, they rarely outperform standard CMLM.
However, sampling masks based on known morpheme boundaries consistently improves performance, highlighting a promising direction for low-resource morphological modeling.
\end{abstract}

\section{Introduction}
Rapid progress in natural language processing (NLP) has largely been driven by training transformer models on massive amounts of unlabeled data.
However, for many of the world’s languages, such large datasets do not exist.
Still there is great potential for NLP to contribute to areas like language documentation or indigenous language educational technology, where both labeled and unlabeled data is extremely sparse, e.g., via character-level tasks like text normalization, grapheme-to-phoneme conversion, or morphological inflection.

Overcoming low-resource data challenges typically involves biasing an NLP model towards a particular language, task or domain.
For instance, in the morphology community, prior work has focused on implementing neural architectures with an inherent inductive bias towards particular tasks like inflection \cite{aharoni2017morphological,makarov-clematide-2018-imitation,wu2019exact}. 

Instead, we take inspiration from high-resource NLP where progress has been largely data-driven, focused on training NLP models with self-supervised objectives like language modeling.
We explore self-supervised objectives, based on masked language modeling \cite[MLM]{devlin-etal-2019-bert}, that have more inductive bias towards a particular character-level task for extremely small datasets in a multitask training setup.
The goal is to change the MLM objective in order to optimally leverage unlabeled data to bias the model towards a particular task.
This is more interpretable and simpler to implement than adjusting the inductive bias of an architecture.
\begin{figure}
    \centering
    \includegraphics[width=0.95\linewidth]{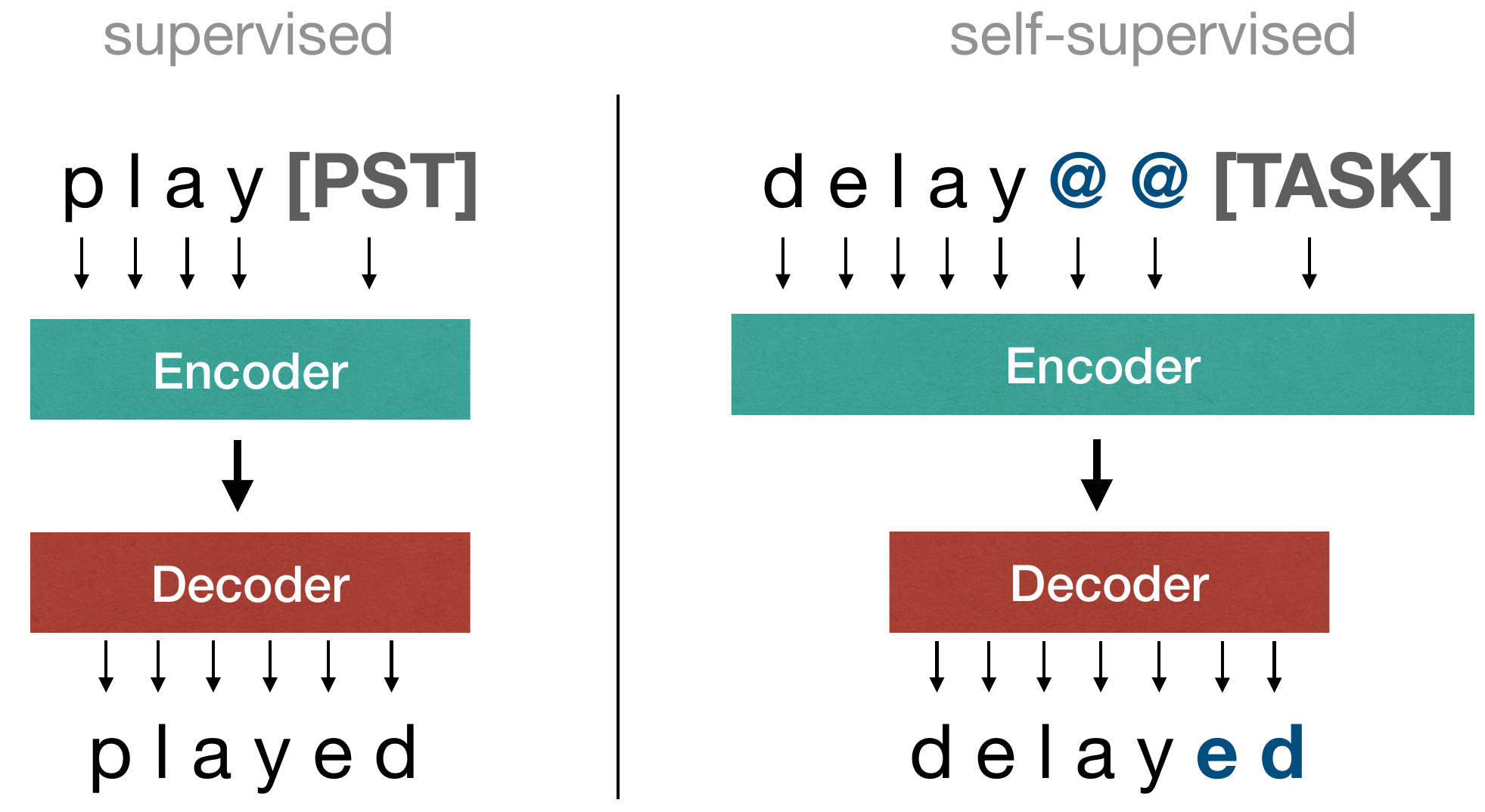}
    \caption{Example of the multitask training; \textit{left:} inflection with the past tense tag; \textit{right:} self-supervised masked language modeling. \textit{@} represents the mask token; \textit{[TASK]} is a special tag for the self-supervised task.}
    \label{fig:multitasking_ex}
\end{figure}

Prior work has shown that auxiliary training on unlabeled data benefits character-level tasks, but has only investigated a small subset of standard objectives.
We hypothesize that this is because 
tasks like MLM at a typical scale are often associated with high computational cost.
We highlight that for many low-resource character-level tasks efficiency is not actually a challenge.
Those tasks allow us to quickly and cheaply explore  variations of auxiliary tasks.
We propose several novel self-supervised task objectives with two goals: (1) to improve downstream task performance, and (2) gain a better understanding of how and why self-supervised auxiliary tasks work.
We focus on morphological inflection, which has been featured in recent work on auxiliary task training \cite{kann-schutze-2017-unlabeled,purushothama-etal-2024-getting}.
Furthermore, it is a character-level task with small models 
that maintains relevance even in the age of LLMs, due to its potential to aid language documentation efforts \cite{moeller-etal-2020-igt2p,moeller-hulden-2018-automatic}.

We conduct experiments with an encoder-decoder transformer \cite{vaswani2017attention} featuring 19 languages, 13 different auxiliary objectives, and 6 different unlabeled datasets.
On a popular inflection benchmark, we find that simply autoencoding unlabeled words, with no masking leads to the best performance when the unlabeled dataset is very small, but MLM performs best as we attain more unlabeled data.
While we find evidence that MLM objectives with more inductive bias 
 influence model predictions in intuitive ways, they typically do not outperform the typical MLM setup on average.
However, we find that sampling masks according to known morpheme boundaries tends to produce the best performance, indicating a promising direction.

\section{Related Work}
\paragraph{Inductive Bias of Auxiliary Tasks}
Following \citet{devlin-etal-2019-bert}, several variations of pretraining tasks have been proposed.
\citet{joshi2020spanbert} propose masking spans of text in order to train models with an inductive bias towards span selection.
\citet{raffel2020exploring} explore pretraining objectives in an encoder-decoder setup and find that a span-denoising objective performs best.
\citet{lewis2020bart} propose additionally shuffling the order of sentences, leading to models particularly strong at machine translation.
More recent work typically assumes a model has been pretrained with autoregressive language modeling, but explore further training like instruction tuning \cite{mishra-etal-2022-cross,wei2021finetuned}, or reinforcement learning \cite{christiano2017deep,ouyang2022training} to align large language models to human preferences.


\paragraph{Auxiliary Tasks for Character-Level Tasks}
Some work has focused specifically on character-level models in the low-resource setting like we do.
\citet{xue2022byt5} pretrain an encoder-decoder following the objective in \citet{raffel2020exploring}, but at the \textit{byte} rather than subword level.
They perform well on many character-level tasks including morphological inflection.
Closely related to our work, \citet{kann-schutze-2017-unlabeled} train inflection models in a mutlitask training setup with autoencoding as an additional objective.
\citet{purushothama-etal-2024-getting} also explore multitask training for inflection, comparing autoencoding to denoising, and find large performance gains even when unlabeled data is sampled exclusively from the supervised training set.
This implies that models benefit from the task itself, rather than simply the additional data distribution.
Our work builds closely upon this idea.
Whereas \cite{purushothama-etal-2024-getting} shows that the self-supervised objective is impactful beyond the incorporation of new data, we explore which self-supervised objective---among several heuristics---is best on various datasets comprising different properties.
Other work has specifically investigated the inductive bias of autoencoding for inflection as beneficial for \textit{copy bias} \cite{liu-hulden-2022-transformer,yang-etal-2022-generalizing}.

\section{Morphological Inflection}
Morphological inflection is a character-level sequence-to-sequence task.
Such tasks are characterized by small vocabularies $\Sigma$ as well as short source and target strings.
Given a source lemma $\ell \in \Sigma^{*}$, target word $y \in \Sigma^{*}$, and a bundle of morphological feature tags $t \in \tau$, the goal of morphological inflection is to learn a mapping
\label{eq:task}\begin{equation*}
    f(\ell, t) \rightarrow y,
\end{equation*}

\noindent for instance, an inflection example for generating the past tense of the verb \textit{try} would look like:
\begin{align*}
    f(\textit{try}, \text{PST}) \rightarrow \textit{tried}.
\end{align*}

\section{Self-Supervised Training Heuristics}
We compare several heuristics for self-supervised training, which are optimized in a multitask learning setup along with inflection.
Each heuristic is implemented as an MLM method, which we vary in order to change the inductive bias of the learning problem.
We assume a small lexicon $\mathcal{L}$ of words, and an even smaller supervised inflection dataset $\mathcal{D}$ to model with an encoder-decoder.

\paragraph{Baselines}
We compare all heuristics to two baselines.
First, we train exclusively on $\mathcal{D}$, ignoring $\mathcal{L}$ entirely.
We refer to this model as \texttt{baseline}.
We also implement an auxiliary baseline, which in addition to the supervised task, trains on $\mathcal{L}$ with autoencoding (\texttt{AE}).
A self-supervised learning sample with autoencoding would be presented to the model, for instance, as:
\begin{align}
\textit{tried} + \text{[TASK]} \rightarrow \textit{tried},
\end{align}
\noindent where [TASK] is a special symbol for the self-supervised task that takes the role of the inflection tags in the supervised task.
Each heuristic noising method combines an \textit{objective} with a sampling \textit{strategy}, and either \textit{masks} or \textit{deletes} characters.

\paragraph{Objective}
We compare two objectives for MLM.
The first, CMLM \cite{wiemerslage2023investigation}, 
 follows the MLM method from \citet{roberta-arxiv}---which samples 15\% of tokens dynamically and 80\% of the time replaces them with a special $\langle{\text{MASK}}\rangle$ symbol, 10\% replaces them with another token from the vocabulary, or 10\% leaves the token as is---except it is performed at the character level and the mask sampling rate is increased to 25\%.
For instance, representing the mask token with @:
\begin{align}
\textit{tr@@ed} + \text{[TASK]} \rightarrow \textit{tried}
\end{align}
\indent We additionally implement the span based variant for the T5 objective from \citet{raffel2020exploring}.
This differs from CMLM in that the sampled token is \textit{always} replaced with a mask, adjacent masks are merged together into a single mask, and each mask token is unique.
For instance
\begin{align}
\label{eq:t5_source}
\textit{t<X>e<Y>} + \text{[TASK]}
\rightarrow \textit{tried},
\end{align}
where \textit{<X>} and \textit{<Y>} are mask tokens.
Notice that this differs from the details of \citet{raffel2020exploring}, who additionally mask the output sequence, keeping only the dropped out tokens intact:
\begin{align}
\label{eq:t5}
\textit{t<X>e<Y>} + \text{[TASK]} \rightarrow \textit{<X>ri<Y>d}.
\end{align}
In initial experiments, we found that this method in \autoref{eq:t5} severely underperformed masking only the source side as in \autoref{eq:t5_source}, so we exclude it from the experiments presented in this work.

\begin{table}[t!]
\centering
\small
\begin{tabular}{l|llll}
\toprule
Dataset & Samples & Types & Med Len & N-grams \\
\toprule
ud-1k & 5836.26 & 3260.68 & 12.66 & 5725.79  \\
\midrule
ud-200 & 5278.37 & 2659.11 & 11.74 & 4839.58 \\
ud-vnadj & 5073.00 & 3297.58 & 11.63 & 4744.63 \\
ud-wl & 5263.68 & 3155.84 & 12.79 & 5479.89 \\
ud-vnadj-NR & 4976.74 & 4902.68 & 12.24 & 6273.21 \\
ud-wl-NR & 5132.74 & 5059.74 & 12.92 & 7562.37 \\
\bottomrule
\end{tabular}
\caption{Statistics for all datasets, averaged over all languages. This includes the supervised and unlabeled data. Types is the number of unique words, Med Len is the median number of characters in each word, and N-grams is the count of all unique char. bi and trigrams.}
\label{tab:data_stats}
\end{table}

\begin{table}[t!]
\centering
\small
\begin{tabular}{l|llll}
\toprule
Dataset & Samples & Types & Med Len & N-grams \\
\toprule
ud-1k & 6000.0 & 3139.2 & 11.7 & 4779.8 \\
\midrule
ud-200 & 5360.0 & 2475.4 & 10.5 & 4136.0  \\
ud-vnadj & 5360.0 & 3264.6 & 11.8 & 3861.6  \\
ud-wl & 5360.0 & 3013.4 & 11.5 & 4494.6  \\
ud-wl-NR & 5360.0 & 5320.4 & 13.2 & 6829.6  \\
ud-vnadj-NR & 5360.0 & 5320.0 & 13.1 & 5467.0  \\
\midrule
seg & 5360.0 & 5320.4 & 13.3 & 5645.0  \\
\bottomrule
\end{tabular}
\caption{Statistics for all datasets, averaged over the 5 languages with morphological segmentation. This includes the supervised and external data.}
\label{tab:data_stats_seg_languages}
\end{table}

\paragraph{Strategy}
We compare three masking strategies: \texttt{iid}, \texttt{suffix}, and \texttt{prefix}.
For each strategy, we sample 25\% of character positions in a word.
Each strategy varies in the distribution over characters from which we sample characters to be masked.
\texttt{\textbf{iid}} sampling is implemented with a uniform distribution over all characters.
The \texttt{\textbf{suffix}} strategy explores the hypothesis that masking with a bias towards the end of a string should create training samples that resemble concatenative, suffixing inflection, which is typologically pervasive.
Here, we sample from a distribution that is skewed toward the end of the sequence.
We distribute 95\% of the probability mass uniformly over the final 1/3 of characters, and the other 5\% uniformly over the first 2/3 of characters.
Finally, for the \texttt{\textbf{prefix}} strategy, which is typologically less common for inflection and not well-represented in our dataset, we do the same thing as the suffix strategy, but skew the distribution to the start of the sequence.

\paragraph{Character Deletion}
We additionally explore deleting rather than masking characters.
Intuitively, given that inflection is a sequence-to-sequence task where in typical concatenative inflection some substring is appended to the input sequence, deleting characters of a word and then predicting them simulates the target task more closely.
For instance, if we sample the word `baked', and apply the \texttt{suffix} strategy, we might produce the training sample
\begin{align}
\textit{bake} \rightarrow \textit{baked}.
\end{align}
which exactly resembles a supervised sample for producing the past tense without the inflection tag.
Note that, with deletion, T5 and CMLM result in \textit{almost} the same denoising method, since merging masks into spans is irrelevant.
The only difference is that CMLM sometimes masks with another character in the vocabulary (10\%) or applies the identity function to a character sampled for masking (10\%).

\begin{figure*}
    \centering
    \includegraphics[width=0.8\linewidth]{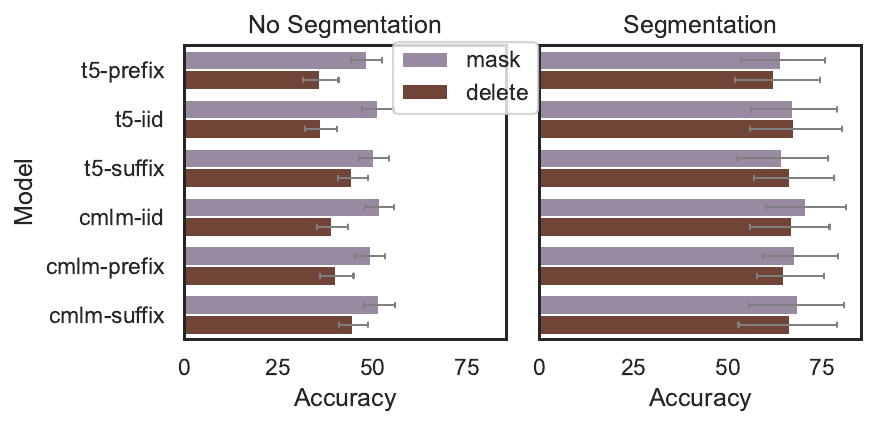}
    \caption{Results for masking v.s. deleting for all models averaged over all datasets. On the right are the segmentation variants of all models, for which there is only one dataset comprising only 5 languages.}
    \label{fig:delete_v_nodelete_all}
\end{figure*}
\paragraph{Segment Masking}
We further explore if masking subword segments according to oracle morpheme boundaries results in an objective more suitable for morphological inflection.
For a subset of languages, we attain gold morpheme boundaries for the unlabeled words, and query this as an oracle.
For this dataset, we produce self-supervised data that samples segments, rather than characters, for masking.
All details are otherwise identical to character-level masking.
For instance, given the word `walk~-~ing', the CMLM objective and suffix strategy, we will produce with 95\% probability
\begin{align}
\textit{walk@@@} \rightarrow \textit{walking},
\end{align}
where the segment -ing is masked.

\section{Experiments}

\begin{table*}[t!]
\centering
\small
\begin{tabular}{lllllllll}
\toprule
  & \texttt{baseline} & \texttt{t5-pref} & \texttt{cmlm-pref} & \texttt{t5-suff} & \texttt{cmlm-suff} & \texttt{t5-iid} & \texttt{cmlm-iid} & \texttt{AE} \\
\midrule
ud-1k & 64.39* & 73.4 & 69.91 & 73.43 & 74.07 & 74.39 & \underline{74.67} & \textbf{75.83} \\
\midrule
ud-200 & 5.16 & 38.31 & 41.82 & 41.34 & 42.76 & 41.04 & \underline{42.92} & \textbf{47.48} \\
ud-wl & 5.16 & 38.91 & 43.16 & 43.07 & \underline{44.98} & 43.05 & 43.98 & \textbf{49.55} \\
ud-vnadj & 5.16 & 39.51 & 41.46 & 41.86 & \underline{43.39} & 42.86 & 42.91 & \textbf{47.8} \\
\midrule
ud-wl-NR & 5.16 & 48.04 & 48.76 & 49.19 & 50.51 & \underline{51.26} & \textbf{51.68} & 50.49 \\
ud-vnadj-NR & 5.16 & 47.67 & 47.88 & 49.2 & 49.42 & \textbf{51.48} & \underline{50.46} & 50.06 \\
\bottomrule
\end{tabular}
\caption{Accuracy for all main experiments averaged over all languages. Note that baseline results are the same because each dataset shares the exact same supervised training samples. The best performing self-supervised setup per dataset is \textbf{bolded}, and the 2nd best is \underline{underlined}.
*Taken from published results in \citet{purushothama-etal-2024-getting}}
\label{tab:mean_results}
\end{table*}

\subsection{Data}
We follow \citet{purushothama-etal-2024-getting} in evaluating on SIGMORPHON 2023 shared task data \cite{goldman2023sigmorphon} with a subset of training data, and unlabeled data sampled from universal dependencies (UD) treebanks \cite{ud-2.12}.
First, we produce results on the exact dataset from \citet{purushothama-etal-2024-getting}, comprising 19 languages, 1k supervised samples, and 5k unlabeled words, which we refer to as \textbf{ud-1k}.
The 19 languages are the subset of SIGMORPHON 2023 data for which UD corpora are also available.
In all other experiments, we simulate a lower-resource setting than that work.
For each language we sample 200 supervised samples from the SIGMORPHON 2023 training set for each part of speech.
This means that each dataset contains 200 -- 600 supervised samples depending on which parts of speech are present in the shared task data (nouns, adjectives, or verbs).
Sampling is done one paradigm at a time, ensuring a broad variety of inflection tags are attested.
This results in supervised datasets similar in size to the low-resource setting studied in the SIGMORPHON 2022 shared task on inflection \cite{kodner-etal-2022-sigmorphon}, which sampled 700 inflection pairs.
The upper bound on our supervised datasets is 600 inflection pairs if all three of verbs, nouns, and adjectives are available.
We augment these supervised samples with 5k unlabeled samples from UD.
The unlabeled data is either copied from \citet{purushothama-etal-2024-getting} (\textbf{ud-200}), or resampled with additional constraints.
In one dataset we constrain word-length: ensuring that all unlabeled words contain at least three characters (two for Japanese) to increase the chance that the dataset contains morphologically rich words (\textbf{ud-wl}).
We build another dataset that samples from words that are additionally tagged with a verb, noun, or adjective part of speech in the UD treebank, to further increase the likelihood of morphological productivity (\textbf{ud-vnadj}).

All unlabeled data so far has been sampled \textit{with} replacement.
Given the size of the UD corpora, this leads to several languages with many duplicate words in the 5k unlabeled samples (See \autoref{tab:data_stats}).
We thus also sample datasets without replacement (\textbf{ud-wl-NR}, \textbf{ud-vnadj-NR}) to increase the diversity of the dataset, leading to 5k unique words in most cases.
This means that the \textbf{*-NR} datasets effectively increase the size of the unlabeled dataset.
We sample from additional UD corpora where necessary in order to ensure that we can sample 5k unique words.
This was not possible for Amharic (amh) nor Sanskrit (san), so those datasets continue to contain fewer than 5k words.
Finally, we build a dataset where all unlabeled words are segmented into morphemes (\textbf{seg}).
For this, we use the data from the SIGMORPHON 2022 morpheme segmentation shared task \cite{batsuren-etal-2022-sigmorphon}.
This dataset contains \textit{canonical} segmentation, but we require a \textit{surface} segmentation.
We produce surface segmentation boundaries with a heuristic algorithm presented in \autoref{sec:seg_algo_appendix}.
Because the overlap in languages between the two shared tasks is small, this dataset contains only 5 languages: English, Hungarian, Italian, Russian, and Spanish.
This dataset is very morphologically rich, because the shared task sampled words contain many morphemes.
Statistics in \autoref{tab:data_stats_seg_languages} show a high median word length, and a large number of unique character n-grams.
The notably high number of n-grams in the ud-wl-NR dataset is likely due to noisy web-scraped data---which is pervasive in the English data---that results in rare characters and character n-grams.
We expect the seg dataset to represent many inflection variations, but also many non-inflectional processes, like derivation, or, e.g., Spanish pronominal participles.

\subsection{Model}
For all experiments, we train an encoder-decoder transformer \cite{vaswani2017attention} following \citet{wu-etal-2021-applying}.
We use a multitask training setup, where both the supervised and self-supervised task can occur in the same batch, and their losses are summed.
Inflection tags are included as additional tokens in the input sequence.
For the self-supervised task, these are replaced with a special task tag.
Exact hyperparamters are available in \autoref{sec:hyperparameters}.
Each model is 7.4 million parameters, and trains in about 1 hour on an NVIDIA A100.

\section{Results}
We first investigate the impact of masking vs. deleting characters for the denoising objectives.
In \autoref{fig:delete_v_nodelete_all} masking outperforms deleting for all training setups on average over all datasets.
However, when the training heuristic has knowledge of morpheme boundaries, this is not strictly true.
In this case, for both \texttt{t5-suffix} and \texttt{t5-iid} deletion is stronger.
In all subsequent experiments we focus on masking, which is typically the superior strategy.

Next we present the main results in \autoref{tab:mean_results} averaged over all languages.
For per-language results see the Appendix.
We first reproduce the results from \citet[ud-1k]{purushothama-etal-2024-getting} showing that autoencoding outperforms all denoising methods.
The same denoising method evaluated in their work, \texttt{cmlm-iid}, is the best performer after autoencoding.
Then we evaluate in our setting, which constrains the supervised data to a smaller training set, while maintaining the same UD dataset for self-supervision.
Here, we attain the same relative result but the gap in performance between autoencoding and the best denoising setup is much larger (4.56 absolute accuracy), due to the smaller supervised dataset.
When we sample the unlabeled UD data with a minimum word-length constraint (ud-wl), we see an increase in accuracy across the board, though autoencoding still performs best by a large margin.
Here, however, the suffix strategy is the best denoising method.
Additionally constraining data to only verbs, adjectives and nouns (ud-vnadj) actually reduces performance when compared to the ud-wl dataset.
This is likely because it reduces the diversity of the dataset due to those constraints (See: \autoref{tab:data_stats}, N-grams).

\paragraph{Sampling Words without Replacement}
\autoref{tab:data_stats} shows that all datasets so far have several fewer unique types than the total number of samples.
This is because we followed \citet{purushothama-etal-2024-getting} by sampling words from UD with replacement, causing many duplicate words.
When we resample the datasets without replacement, effectively increasing the size of the unlabeled datasets to be truly 5000 samples, there is an increase in accuracy across the board, with a substantial increase for all denoising strategies  (up to 9.13 absolute accuracy).
For these datasets, denoising clearly outperforms autoencoding.

\begin{figure*}[t!]
    \centering
    \begin{subfigure}[t]{0.5\textwidth}
        \centering
        \includegraphics[width=0.9\linewidth]{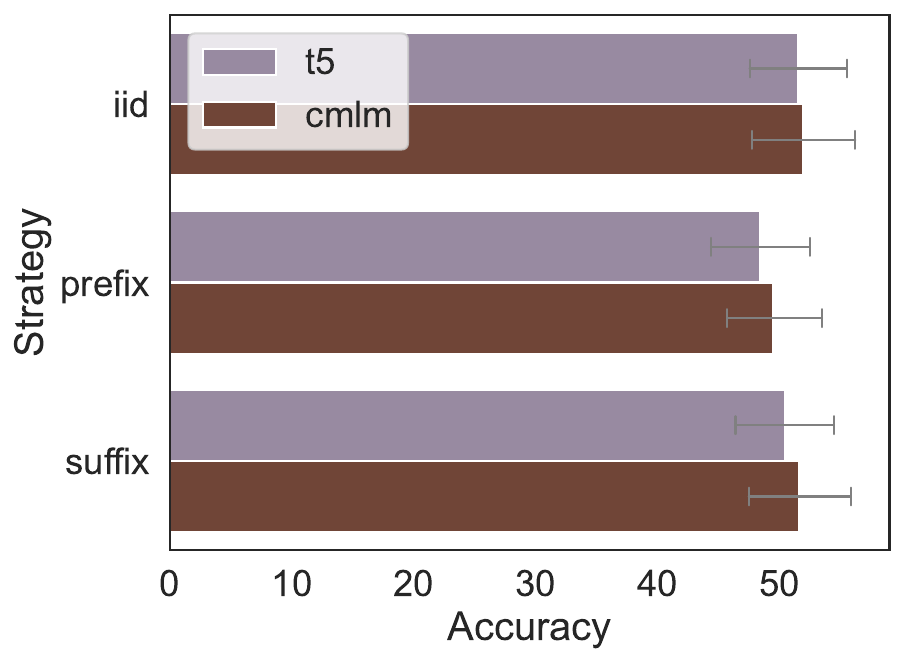}
        \caption{Comparison of objective performance}
        \label{fig:ob_sampling}
    \end{subfigure}%
    ~ 
    \begin{subfigure}[t]{0.5\textwidth}
        \centering
        \includegraphics[width=0.9\linewidth]{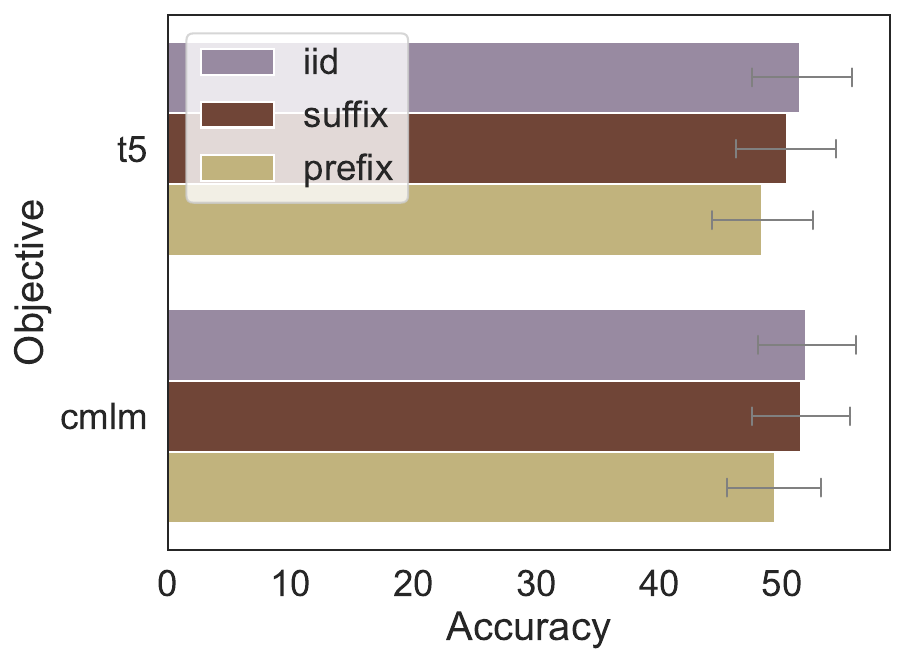}
        \caption{Comparison of strategy performance}
        \label{fig:strat_sampling}
    \end{subfigure}
    \caption{Average accuracy over all ud datasets when masking.}
    
\end{figure*}

\paragraph{Objective}
\autoref{fig:ob_sampling} presents a comparison of the two self-supervised objectives averaged over all languages and datasets on the left.
For all strategies, CMLM outperforms T5 on average.
Despite this, when combined with the best strategy (\texttt{iid}), the difference of these averages is rather small.

\paragraph{Strategy}
\autoref{fig:strat_sampling}  presents a comparison of the three self-supervised strategies averaged over all languages and datasets.
The \texttt{iid} strategy performs best on average for both objectives, followed closely by the \texttt{suffix} strategy.

\begin{table*}[t!]
\centering
\small
\begin{tabular}{lllllll}
\toprule
 & eng & hun & ita & rus & spa & mean \\
\midrule
\texttt{AE} & \underline{90.70} & 52.00 & \underline{53.70} & 68.40 & 55.20 & 64.00 \\
\midrule
\texttt{t5-pref} & 88.70 & 58.60 & 46.90 & 69.80 & 63.30 & 65.46 \\
\texttt{t5-suff} & 89.40 & 66.60 & 50.20 & 67.70 & 57.90 & 66.36 \\
\texttt{t5-iid} & 90.10 & \textbf{68.80} & 47.90 & 70.40 & \textbf{68.00} & 69.04 \\
\texttt{cmlm-pref} & 88.70 & 56.90 & 42.10 & 69.50 & 59.90 & 63.42 \\
\texttt{cmlm-suff} & 90.50 & 66.70 & 51.90 & 68.50 & 61.00 & 67.72 \\
\texttt{cmlm-iid} & 90.60 & 67.30 & 52.60 & \underline{73.10} & 66.30 & \underline{69.98} \\
\midrule
\texttt{t5-seg-pref} & 86.60 & 58.00 & 46.70 & 67.60 & 60.70 & 63.92 \\
\texttt{t5-seg-suff} & 89.50 & 58.40 & 45.30 & 71.60 & 56.10 & 64.18 \\
\texttt{t5-seg-iid} & 89.10 & 63.70 & 48.70 & 71.60 & 62.50 & 67.12 \\
\texttt{cmlm-seg-pref} & 89.70 & 57.60 & \textbf{58.20} & 68.60 & 64.20 & 67.66 \\
\texttt{cmlm-seg-suff} & \textbf{91.00} & 67.00 & 47.90 & \textbf{73.20} & 64.10 & 68.64 \\
\texttt{cmlm-seg-iid} & 90.50 & \underline{68.30} & 53.20 & \textbf{73.20} & \underline{67.60} & \textbf{70.56} \\
\bottomrule
\end{tabular}
\caption{Accuracy for the languages with segmentation data. -seg- indicates the masking objective sampled morphemes, rather than characters. The best model per language is \textbf{bolded}, and the second best is \underline{underlined}}
\label{tab:seg_results}
\end{table*}

\paragraph{Segmentation}
Finally, we present results for the five segmentation languages in \autoref{tab:seg_results}.
When using the CMLM objective, the segmentation variant---which samples entire morphemes for masking rather than individual characters---always outperforms the corresponding setup without segmentation.
However, the opposite is true for T5 -- in these cases the segmentation variant \textit{under-performs}.
\autoref{fig:delete_v_nodelete_all} also shows that deleting tends to outperform masking for the T5 objective at the segment level.
The best performing system in this experiment on average is \texttt{cmlm-seg-iid}.
However, the same setup without segmentation performs only 0.58 absolute accuracy lower on average.

\begin{figure*}
    \centering
    \includegraphics[width=\linewidth]{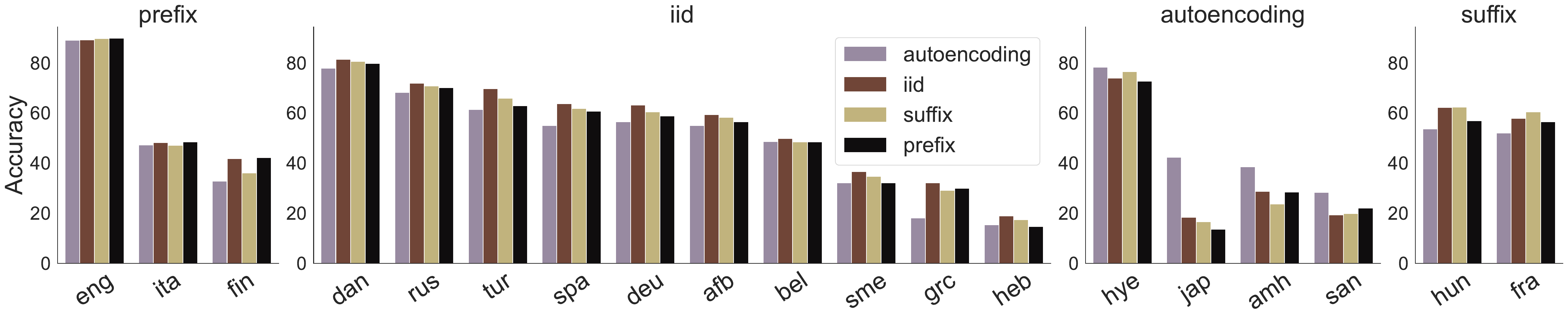}
    \caption{Language specific results for the ud-vnadj-NR dataset. Bars represent the max test accuracy of each strategy. Languages are grouped by which strategy maximized performance on this dataset.}
    \label{fig:lang_spec}
\end{figure*}

\subsection{Language-Specific Results}
\label{sec:language_spec_results}
We perform a small language-specific analysis focused on the self-supervised strategy, which we expect to be most relevant to language typology.
We focus on the ud-vnadj-NR dataset, which we had expected to contain the most morphologically rich data of our full 19-language datasets.
In \autoref{fig:lang_spec} we present the best performing variation for each strategy on this dataset.
For almost all languages, the \texttt{iid} masking strategy performs best.
There are 4 languages for which \texttt{AE} outperforms every denoising strategy. In three of these four languages either we sampled far fewer than 5k self-supervised words (amh, san), or the character vocabulary is extremely large (jap, amh).
Both of these scenarios support the hypothesis that autoencoding is a strong objective under extreme data sparsity.
It is unclear which factor explains the autoencoding performance in Armenian (hye).
Despite the fact that our dataset contains no languages with typical prefix inflection, the \texttt{prefix} strategy also performs best in a few cases.
Upon further analysis, we noticed that in the development sets, particles like pronouns are sometimes inserted before an inflected form as in the Italian example:
\begin{align}
\textit{sdebitarsi} \rightarrow \textit{ti sdebitavi}.
\end{align}
When we quantify this phenomenon, we find that in 1000 development set examples, it happens most frequently in fin (284) and ita (141), followed by hye (99) and deu (71).
The German (deu) instances are caused by separable verbs, and Armenian (hye) additionally exhibits this phenomenon pervasively at the end of the string.
We interpret this as evidence that skewing the mask distribution 
does teach the model an inductive bias towards certain typological phenomenon.
The fact that many languages that exhibit a strong bias towards suffixes perform best with the \texttt{iid} strategy, however, shows that masking uniformly is typically stronger overall.

\subsection{Discussion}
The main results presented in this work show a clear trend between autoencoding---which performs strongly when there is little external unlabeled data---and denoising, which performs strongly as we train on a larger set of unlabeled words.
This is apparent in the trend in the number of unique types per dataset in \autoref{tab:data_stats} vis-à-vis the accuracy in \autoref{tab:mean_results}.
We hypothesize that this is because autoencoding learns a strong inductive bias towards \textit{copying} -- a common operation in morphological inflection.
On the other hand, denoising encourages models to generate particular character sequences that were masked in the input.
It is thus prone to overfitting to such sequences in cases of extreme data sparsity.
Denoising also encourages the model to learn to generate character sequences that are potentially morphologically rich and unavailable in the source form in an inflection (e.g., an inflectional suffix).
A morphological inflection model must learn to do this competently, and with sufficiently diverse data, we hypothesize that denoising objectives encourage this capability.
We are not the first to hypothesize that autoencoding benefits inflection via copy bias.
\citet{liu-hulden-2022-transformer} proposed generating synthetic words or lemmas and training inflection models to copy them. \citet{yang-etal-2022-generalizing} confirmed that this strategy strongly increases inflection performance.

\begin{figure*}[t!]
    \centering
    \begin{subfigure}[t]{0.45\textwidth}
        \centering
        \includegraphics[width=0.9\linewidth]{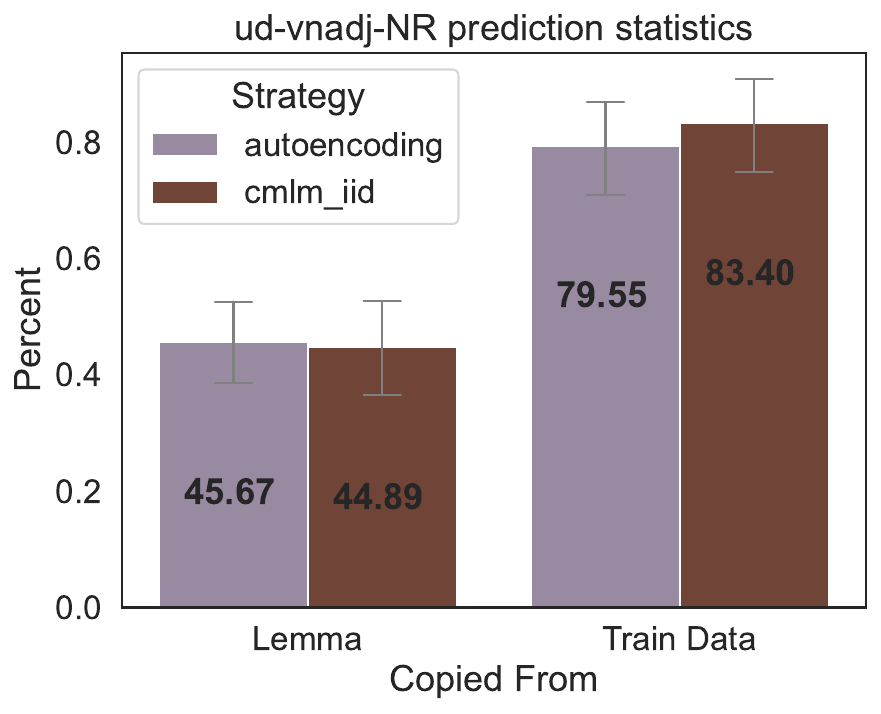}
    \end{subfigure}%
    ~ 
    \begin{subfigure}[t]{0.45\textwidth}
        \centering
        \includegraphics[width=0.85\linewidth]{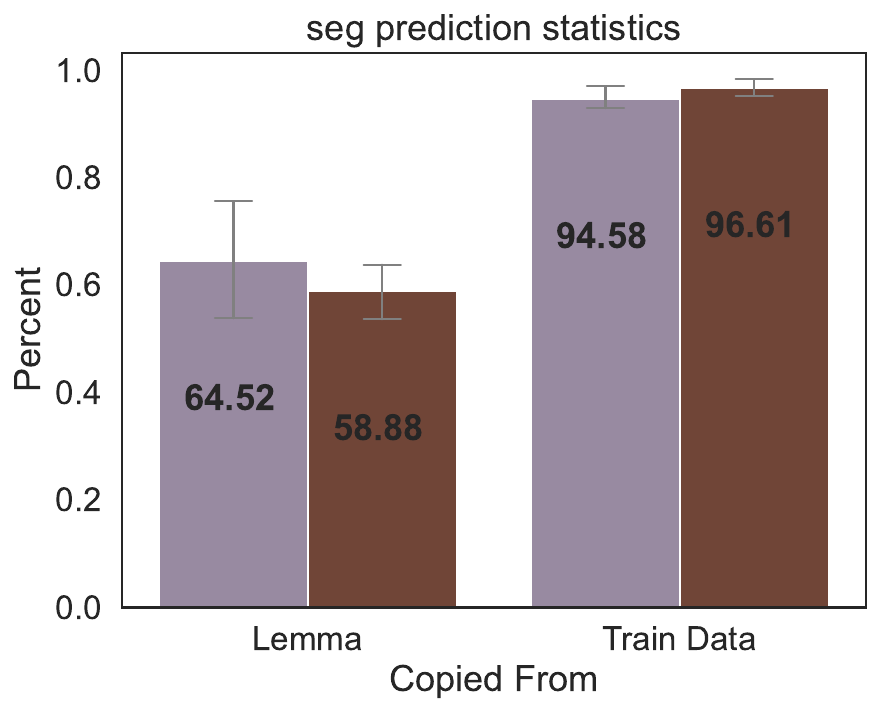}
    \end{subfigure}
    \caption{The percent of trigams in predicted inflections that were copied from the lemma or from the training outputs. This compares models trained with autoencoding to \texttt{cmlm-iid} with masking on two different datasets.}
    \label{fig:pred_stats}
\end{figure*}

However, if we consider \texttt{cmlm-iid} with masking as a baseline denoising method, all of our attempts to change the objective with a different inductive bias can be considered negative results.
Deleting, merging span masks, and skewing the positions of the mask distribution do not lead to better results overall.
One hypothesis here is simply that morphology is complicated in most languages: even if suffix concatenation, for instance, is pervasive, it does not account for all of the inflectional transformations and is thus too strong of an inductive bias.
Instead, we believe that \texttt{iid} sampling with masks is the best heuristic because it leads to the highest variance in training data.
That is, given a small unlabeled dataset, we sample each word many times during training.
A uniform sampling of masks will lead to the highest variance in word positions that are masked for the same word, and thus the most variance in our dataset.
This is likely a stronger learning method than any particular inductive bias.

In \autoref{sec:language_spec_results} we presented evidence that \texttt{prefix} masking is beneficial to inflection datasets with pervasive prefix inflection.
This suggests that the auxiliary self-supervised task \textit{does} modify model behavior in typologically interpretable ways.
However, the relatively weak \texttt{suffix} results when compared to \texttt{iid} suggests that inductive bias is less beneficial than simply learning the most diverse token distribution, even in a low-resource setting.

\paragraph{Segment Masking}
Masking morphemes rather than characters is more closely aligned to the downstream inflection task for concatenative inflections.
In our experiments, we tend to see a slight positive effect from incorporating morpheme information to this end.
This indicates that smarter masking \textit{can} increase performance, though on average sampling characters uniformly is a very strong alternative.
We observe two other results that are unique to the seg dataset: (i) morpheme masking \textit{underperforms} character masking with the T5 objective (ii) deleting sometimes outperforms masking.
The first result may be due to the fact that T5 on the seg dataset is most likely to have a large mismatch between the number of symbols on the source and target side.
That is, an entire segment comprising many characters is replaced by a single mask symbol.
For instance, if we sample the stem in the following example:
$\textit{<X>s} + \text{[TASK]}
\rightarrow \textit{recovers}$.
Instead, the CMLM objective would result in seven individual mask tokens that align with the stem.
The fact that deletion still performs strongly for the T5 objective with segments (See: \autoref{fig:delete_v_nodelete_all}), however, may warrant further investigation into the impact of masks.
See \autoref{tab:seg-langs-delete} in the appendix for full results on the seg dataset with deletion.

\section{Additional Analysis: Copying}
In order to explore our hypothesis that autoencoding teaches the model to copy, whereas denoising teaches the model to generate from the training distribution, we quantify the extent to which models copy n-grams from the lemma, versus generate n-grams from the training data.
To this end, we analyze predictions on the development set for the ud-vnadj-NR dataset used in the language-specific analysis.
We compare the autoencoding model, and the \texttt{cmlm-iid} mask model.
For each prediction, we compute the percent of character trigrams that appear in the trigrams of the input lemma in order to quantify copying.
Then, we build the set of all character trigrams in the training data output forms, and compute the percent of predicted character trigrams that are attested in the training trigrams in order to quantify generating from the training distribution.
In \autoref{fig:pred_stats} we compare the average percent of the two models over all languages for those predictions where the models disagree.
This represents 7286 total disagreements for ud-vnadj-NR and 2511 for the seg dataset.
In both datasets, autoencoding tends to copy more from the lemma and denoising tends to produce more trigrams from the training data.
We interpret this as support for our hypothesis.
In the segmentation dataset, the difference in copy percent is much more substantial, and the training data percent difference is much smaller.
The latter effect is likely because this morphologically rich dataset contains most n-grams that a model is likely to generate.
The former effect illustrates that many \texttt{AE} mistakes probably come from over-copying.

\section{Conclusion}
In this work, we compared several self-supervised tasks performed on unlabeled data as auxiliary training objectives for morphological inflection.
We found that for small sets of unlabeled words, autoencoding is a strong objective due to its inductive bias towards copying form the source string.
However, tasks based on MLM perform better as the unlabeled dataset grows because it encourages the model to generate new character sequences.
Other denoising objectives with more inductive bias have predictable effects, but are not necessarily better than the standard MLM with uniform masking. 
We believe this is because uniform masking produces the most diverse dataset.
Finally, we find that providing a morpheme segment oracle leads to the best results, establishing a promising direction if segments can be approximated without supervision.

\section*{Limitations}
First, this study is limited with respect to the morphological typology of the languages.
Because we use UD data, we filter out the SIGMORPHON 2023 shared task languages that tend to inflect with prefixes, for instance.

We also explore a limited set of simple self-supervised heuristics.
Other methods may be more effective for inflection.
Our segmentation dataset contains strictly concatenative languages, and it is unclear what this means for non-concatenative morphology, which tends to be understudied in the computational morphology community.

Finally, we claim that better inflection models can be useful for language documentation efforts, but we test all methods on a standard benchmark by \textit{simulating} a low-resource scenario.
Languages with data needs like those that we discuss may yield different results.

\section*{Ethics Statement}
This work is partly inspired by applications of language technology to language documentation and educational tools.
We note that many of these applications traditionally apply to small indigenous language communities, who may not actually desire or need such NLP technology.
We thus emphasize that the methods proposed here are intended to explore the development of NLP tools under severe data constraints, but do not test the feasibility of using them in a real applied setting.

\section*{Acknowledgments}
We would like to thank Abteen Ebrahimi, Stéphane Aroca-Ouellette, Ananya Ganesh, and Kyle Gorman for feedback on this work.
We would also like to thank Kyle Gorman for help with the alignment algorithm for turning canonical segmentation into surface segmentation.

\bibliography{anthology,custom}
\bibliographystyle{acl_natbib}

\newpage

\appendix

\section*{Appendix}
\label{sec:appendix}

\begin{table*}[t!]
\centering
\begin{tabular}{llllllll}
\toprule
 & \texttt{t5-prefix} & \texttt{cmlm-prefix} & \texttt{t5-suffix} & \texttt{cmlm-suffix} & \texttt{t5-iid} & \texttt{cmlm-iid} & \texttt{AE} \\
\midrule
afb & \textbf{75.80} & \underline{74.50} & 71.90 & 73.70 & 74.40 & 73.30 & 72.60 \\
amh & 59.90 & 62.80 & 61.80 & 62.00 & 60.60 & \underline{63.80} & \textbf{69.20} \\
bel & \underline{64.00} & 63.30 & 62.40 & 62.40 & 62.40 & \textbf{64.90} & 62.20 \\
dan & 81.70 & 82.20 & 82.90 & \textbf{83.90} & \underline{83.50} & 82.70 & 82.30 \\
deu & 74.10 & \underline{77.60} & 73.80 & 77.50 & 76.10 & \textbf{79.00} & 76.90 \\
eng & 88.10 & \textbf{90.50} & 89.00 & 89.50 & 89.40 & 88.50 & \underline{90.10} \\
fin & \underline{73.90} & 72.50 & \textbf{74.60} & 73.20 & \textbf{74.60} & 70.70 & \underline{73.90} \\
fra & 75.80 & 74.10 & \textbf{76.40} & \underline{76.10} & 73.70 & 74.70 & 75.00 \\
grc & 52.50 & 49.60 & 50.80 & 47.90 & \underline{53.40} & \textbf{55.30} & 49.20 \\
heb & 76.94 & 77.04 & \underline{77.84} & 77.24 & 77.64 & \textbf{79.66} & 76.84 \\
hun & 77.70 & 3.40 & 77.90 & 79.50 & \textbf{80.50} & 77.60 & \underline{79.80} \\
hye & 91.30 & \underline{92.40} & 91.60 & 92.20 & \textbf{93.30} & 92.30 & \textbf{93.30} \\
ita & \textbf{92.40} & \underline{92.10} & 91.20 & 90.80 & 92.00 & 91.00 & 91.30 \\
jap & 37.80 & 39.10 & 41.50 & \underline{44.20} & 37.80 & \underline{44.20} & \textbf{53.60} \\
rus & 79.50 & 80.10 & 80.20 & \underline{82.20} & \textbf{82.40} & \underline{82.20} & 81.30 \\
san & 56.40 & 56.10 & 54.80 & 56.80 & \underline{59.10} & 57.10 & \textbf{61.10} \\
sme & 61.00 & 61.30 & 61.10 & 59.60 & \underline{66.20} & 64.50 & \textbf{69.20} \\
spa & 91.20 & 91.60 & 90.30 & \underline{92.40} & 90.60 & 92.30 & \textbf{93.10} \\
tur & 84.60 & \underline{88.00} & 85.10 & 86.10 & 85.70 & 84.90 & \textbf{89.90} \\
Mean & 73.40 & 69.91 & 73.43 & 74.07 & 74.39 & \underline{74.67} & \textbf{75.83} \\
\bottomrule
\end{tabular}
\caption{All language results for the ud-1k dataset from \cite{purushothama-etal-2024-getting}.}
\label{tab:abhishek_paper_results}
\end{table*}

\section{UD Data}
\label{sec:ud_appendix}
When we sample from UD data, we use exactly the corpora from \citet{purushothama-etal-2024-getting} except for two cases.
For both English and Turkish, there is not sufficient data in the existing corpora to sample 5k unique words from, so we additionally add the English-EWT corpus and the Turkish-BOUN, respectively.
The English-EWT contains web data, and so we ignore tokens with special sequences "@" and "www".
This process is imperfect though, so we still introduce noise into our data via this corpus.

\section{Surface Segmentation Data}
\label{sec:seg_algo_appendix}
Canonically segmented data may insert or delete characters into or from a surface word form.
We thus present a simple method for attaining a surface segmentation given a surface form $s$ and its canonical segmentation $c$.
First, we build a Pynini \texttt{EditTransducer} \cite{gorman2016pynini}---which is a weighted finite state transducer aligning strings---over the language's character vocabulary $\Sigma$.
Then, we attain the shortest path from $s$ to $c$: the alignment that produces their minimal Levenshtein distance.
Finally, we segment $s$ (i) at any insertion that maps to a segmentation boundary in $c$ or (ii) after any character that maps to a segmentation boundary in $c$.

For example, given the pair $\langle{\textit{chugged, ~~ chug-ed}}\rangle$, we attain the alignment
\begin{align*}
    &\textit{c ~ h ~ u ~ g ~ g ~ e ~ d} \\
    &\textit{c ~ h ~ u ~ g ~ - ~ e ~ d} \\
\end{align*}
Then we match on rule (ii) and segment $s$ into $\textit{chugg - ed}$.

\section{Hyperparameters}
\label{sec:hyperparameters}
Our architecture hyperparameters match \citet{wu-etal-2021-applying}:
We use a transformer encoder and decoder, each with 4 layers, embedding size 256, hidden size 1024, and 4 attention heads.
We optimize with cross-entropy loss with label smoothing of 0.1.
We train with Adam optimization \cite{kingma2014adam} with a learning rate of 1e-3, and a beta2 of .98.
We train up to 800 epochs with a batch size of 400, and evaluate every 16 epochs. 
For the baseline experiments, we lower the batch size to 100 and train for up to 10k epochs with a patience of 400.
We train with the Warmup Inverse Square Root schedule with 4k warmup steps.
When masking unlabeled data, we always sample 25\% of tokens for masking.
All experiments are trained on NVIDIA GPUs and implemented with yoyodyne\footnote{\url{https://github.com/CUNY-CL/yoyodyne}} \cite{wiemerslage-etal-2024-quantifying}.

\begin{table*}[t!]
\centering
\begin{tabular}{llllllll}
\toprule
 & \texttt{t5-prefix} & \texttt{cmlm-prefix} & \texttt{t5-suffix} & \texttt{cmlm-suffix} & \texttt{t5-iid} & \texttt{cmlm-iid} & \texttt{AE} \\
\midrule
afb & 54.30 & 53.30 & 54.30 & 53.10 & \textbf{56.50} & 52.90 & \underline{54.40} \\
amh & 26.00 & \underline{29.30} & 26.40 & \underline{29.30} & 26.00 & 28.40 & \textbf{37.10} \\
bel & 47.00 & \textbf{50.60} & 45.20 & \underline{48.60} & 45.20 & \underline{48.60} & 44.90 \\
dan & 79.00 & \textbf{80.10} & 79.80 & 79.60 & \underline{79.90} & 79.60 & 79.40 \\
deu & 51.40 & 55.60 & 54.40 & 57.90 & \underline{58.50} & \textbf{60.50} & 53.90 \\
eng & 28.90 & \underline{51.60} & 50.90 & 46.60 & 35.40 & 48.70 & \textbf{75.40} \\
fin & \textbf{38.70} & 28.40 & 31.60 & 33.30 & \underline{38.50} & 30.00 & 28.70 \\
fra & 40.40 & \underline{47.50} & 46.30 & 45.30 & 46.80 & 45.10 & \textbf{49.70} \\
grc & 24.00 & 19.50 & 22.50 & 26.00 & \textbf{30.70} & \underline{29.20} & 21.60 \\
heb & 14.60 & \textbf{16.01} & 13.29 & 12.49 & \underline{15.11} & 14.60 & 14.30 \\
hun & 49.00 & 55.30 & \textbf{58.70} & 56.60 & 51.80 & \underline{58.40} & 56.90 \\
hye & 60.90 & 65.10 & \textbf{73.10} & 71.40 & 68.90 & 61.10 & \underline{72.40} \\
ita & 27.10 & 37.00 & 34.20 & 40.10 & 33.30 & \underline{41.40} & \textbf{48.60} \\
jap & 0.80 & 1.70 & 2.30 & \underline{3.70} & 2.00 & 2.00 & \textbf{21.40} \\
rus & 66.80 & 66.00 & 63.10 & 66.20 & 66.10 & \textbf{72.00} & \underline{67.40} \\
san & 20.10 & 23.50 & 26.50 & 25.50 & 23.20 & \underline{27.40} & \textbf{33.80} \\
sme & 29.30 & \textbf{33.20} & 28.70 & 28.80 & \underline{32.10} & 28.20 & 31.90 \\
spa & 53.30 & 51.10 & 49.10 & 54.80 & 54.40 & \textbf{57.50} & \underline{56.90} \\
tur & 16.20 & 29.80 & 25.00 & \underline{33.20} & 15.40 & 29.80 & \textbf{53.40} \\
Mean & 38.31 & 41.82 & 41.34 & 42.76 & 41.04 & \underline{42.92} & \textbf{47.48} \\
\bottomrule
\end{tabular}
\caption{All language results for the ud-200 dataset.}
\label{tab:ud-200}
\end{table*}

\begin{table*}[t!]
\centering
\begin{tabular}{llllllll}
\toprule
 & \texttt{t5-prefix} & \texttt{cmlm-prefix} & \texttt{t5-suffix} & \texttt{cmlm-suffix} & \texttt{t5-iid} & \texttt{cmlm-iid} & \texttt{AE} \\
\midrule
afb & 54.90 & 54.20 & \underline{56.20} & 55.70 & \textbf{56.70} & 55.60 & 54.60 \\
amh & 22.40 & 26.30 & 26.30 & 24.30 & \underline{27.40} & 25.00 & \textbf{37.80} \\
bel & 48.80 & \textbf{50.20} & 43.10 & 48.70 & 47.80 & 48.20 & \underline{49.30} \\
dan & 80.20 & 79.40 & 77.30 & 78.80 & \underline{80.60} & \textbf{80.70} & 79.20 \\
deu & 54.10 & 57.90 & 52.90 & \underline{58.10} & 57.90 & \textbf{62.70} & 52.90 \\
eng & 23.30 & 44.80 & 40.30 & \underline{45.80} & 29.60 & 36.80 & \textbf{72.70} \\
fin & 36.80 & 28.70 & 33.70 & 34.00 & \textbf{43.50} & \underline{38.10} & 32.20 \\
fra & 43.80 & 43.80 & 48.40 & 49.60 & \textbf{52.60} & 51.50 & \underline{52.50} \\
grc & 23.30 & 20.70 & 21.60 & 26.10 & \underline{26.30} & \textbf{28.70} & 16.90 \\
heb & 15.41 & \textbf{16.72} & 15.41 & 13.09 & \underline{15.71} & 13.60 & 15.41 \\
hun & 53.20 & 55.80 & 59.80 & \underline{60.10} & \textbf{62.70} & 58.30 & 53.70 \\
hye & 69.00 & 66.20 & 72.20 & \underline{73.40} & 69.10 & 63.40 & \textbf{73.90} \\
ita & 38.80 & 43.10 & 42.10 & 43.70 & 44.00 & \underline{45.30} & \textbf{49.70} \\
jap & 6.90 & 9.30 & \underline{11.10} & 9.20 & 5.40 & 7.90 & \textbf{30.80} \\
rus & 64.30 & \textbf{70.50} & 65.00 & 68.80 & 67.40 & \underline{70.00} & 67.90 \\
san & 14.00 & \underline{21.00} & 15.90 & 17.70 & 18.90 & 20.30 & \textbf{30.50} \\
sme & 31.00 & 31.60 & 30.60 & 28.30 & 32.60 & \underline{33.50} & \textbf{33.70} \\
spa & 52.20 & 45.60 & 55.10 & \underline{56.30} & \textbf{59.70} & 55.40 & 52.20 \\
tur & 18.20 & 22.00 & 28.30 & \underline{32.80} & 16.40 & 20.30 & \textbf{52.20} \\
Mean & 39.51 & 41.46 & 41.86 & \underline{43.39} & 42.86 & 42.91 & \textbf{47.80} \\
\bottomrule
\end{tabular}
\caption{All language results for the ud-vnadj dataset}
\label{tab:vnadj}
\end{table*}

\begin{table*}[t!]
\centering
\begin{tabular}{llllllll}
\toprule
 & \texttt{t5-prefix} & \texttt{cmlm-prefix} & \texttt{t5-suffix} & \texttt{cmlm-suffix} & \texttt{t5-iid} & \texttt{cmlm-iid} & \texttt{AE} \\
\midrule
afb & 54.40 & 53.50 & \textbf{57.10} & 55.50 & \underline{56.30} & 54.10 & 55.20 \\
amh & 25.90 & \underline{34.30} & 32.70 & 33.60 & 32.10 & 34.00 & \textbf{40.70} \\
bel & 44.70 & \textbf{49.50} & 42.30 & 47.90 & 47.50 & 45.40 & \underline{48.10} \\
dan & 77.00 & \textbf{79.70} & 77.80 & 79.00 & 78.90 & 78.60 & \underline{79.20} \\
deu & 52.30 & 56.20 & 56.40 & 56.80 & \underline{60.10} & \textbf{63.40} & \underline{60.10} \\
eng & 40.50 & \underline{58.80} & 49.60 & 52.50 & 37.10 & 50.60 & \textbf{80.00} \\
fin & \textbf{38.40} & 28.50 & 32.60 & 30.30 & \underline{38.10} & 31.60 & 34.40 \\
fra & 40.30 & 45.20 & 51.40 & 53.50 & \textbf{56.60} & \underline{54.10} & 53.70 \\
grc & 25.90 & 25.30 & 25.60 & \textbf{30.30} & 28.60 & \underline{29.10} & 21.90 \\
heb & 15.61 & 15.31 & 14.00 & 13.90 & \underline{16.62} & 13.60 & \textbf{18.53} \\
hun & 50.20 & \underline{54.90} & 53.90 & 54.50 & \textbf{59.30} & 54.10 & 53.80 \\
hye & 61.30 & 64.80 & 70.60 & \textbf{77.00} & 72.20 & 70.20 & \underline{74.70} \\
ita & 32.40 & \underline{45.10} & 41.60 & 45.00 & 38.90 & 43.80 & \textbf{46.60} \\
jap & 3.80 & 5.50 & \underline{9.30} & 9.00 & 8.20 & 7.80 & \textbf{29.00} \\
rus & 66.00 & \underline{69.00} & 65.70 & 66.80 & \textbf{69.40} & 66.80 & 68.10 \\
san & 17.30 & \underline{26.50} & 21.70 & 25.80 & 18.00 & \underline{26.50} & \textbf{33.30} \\
sme & 28.90 & \underline{31.00} & 27.80 & \textbf{32.10} & 26.60 & 30.80 & \underline{31.00} \\
spa & 45.80 & 47.30 & 51.90 & 54.50 & 54.10 & \underline{55.00} & \textbf{57.10} \\
tur & 18.50 & 29.70 & 36.30 & \underline{36.70} & 19.30 & 26.20 & \textbf{56.10} \\
Mean & 38.91 & 43.16 & 43.07 & \underline{44.98} & 43.05 & 43.98 & \textbf{49.55} \\
\bottomrule
\end{tabular}
\caption{All language results for the ud-wl dataset}
\label{tab:word_length}
\end{table*}

\begin{table*}[t!]
\centering
\begin{tabular}{llllllll}
\toprule
 & \texttt{t5-prefix} & \texttt{cmlm-prefix} & \texttt{t5-suffix} & \texttt{cmlm-suffix} & \texttt{t5-iid} & \texttt{cmlm-iid} & \texttt{AE} \\
\midrule
afb & 56.50 & 54.50 & \underline{58.30} & 56.80 & \textbf{59.40} & 55.70 & 55.00 \\
amh & 27.90 & 28.50 & 23.70 & 23.30 & 27.90 & \underline{28.80} & \textbf{38.50} \\
bel & 47.60 & 48.50 & 48.40 & 48.50 & \underline{49.60} & \textbf{49.80} & 48.70 \\
dan & 79.10 & 79.90 & 76.70 & \underline{80.70} & \textbf{81.50} & 80.40 & 77.90 \\
deu & 58.50 & 58.80 & 60.50 & 58.90 & \textbf{63.20} & \underline{62.50} & 56.50 \\
eng & 87.20 & \textbf{89.90} & 87.30 & \underline{89.80} & 89.30 & 88.80 & 89.10 \\
fin & \textbf{42.30} & 35.70 & 36.10 & 35.30 & \underline{41.80} & 38.80 & 32.90 \\
fra & 56.50 & 43.10 & \textbf{60.50} & 51.30 & \underline{57.90} & 53.80 & 52.00 \\
grc & 30.00 & 26.00 & 29.20 & 28.00 & \textbf{32.10} & \underline{31.30} & 18.10 \\
heb & 14.70 & 13.70 & \underline{17.42} & 16.21 & \textbf{18.93} & 12.99 & 15.41 \\
hun & 55.60 & 57.00 & 55.30 & \textbf{62.40} & 61.70 & \underline{62.20} & 53.70 \\
hye & 72.70 & 69.60 & 74.40 & \underline{76.60} & 74.00 & 73.40 & \textbf{78.30} \\
ita & 39.50 & \textbf{48.50} & 46.60 & 47.20 & 46.70 & \underline{48.30} & 47.30 \\
jap & 13.70 & 13.40 & 16.70 & 13.50 & \underline{18.40} & 17.10 & \textbf{42.40} \\
rus & 65.90 & 70.20 & 70.30 & \underline{70.80} & 69.20 & \textbf{71.90} & 68.30 \\
san & 13.10 & \underline{22.10} & 19.90 & 17.10 & 18.60 & 19.30 & \textbf{28.40} \\
sme & 32.10 & 31.80 & 29.90 & \underline{34.70} & 34.20 & \textbf{36.70} & 32.20 \\
spa & 60.70 & 55.60 & 59.30 & 61.80 & \textbf{63.80} & \underline{62.80} & 55.00 \\
tur & 52.20 & 63.00 & 64.20 & \underline{66.00} & \textbf{69.80} & 64.10 & 61.40 \\
Mean & 47.67 & 47.88 & 49.20 & 49.42 & \textbf{51.48} & \underline{50.46} & 50.06 \\
\bottomrule
\end{tabular}
\caption{All language results for the ud-vnadj-NR dataset.}
\label{tab:vnadj-r}
\end{table*}

\begin{table*}[t!]
\centering
\begin{tabular}{llllllll}
\toprule
 & \texttt{t5-prefix} & \texttt{cmlm-prefix} & \texttt{t5-suffix} & \texttt{cmlm-suffix} & \texttt{t5-iid} & \texttt{cmlm-iid} & \texttt{AE} \\
\midrule
afb & 55.80 & 52.70 & \textbf{56.60} & 56.00 & \underline{56.40} & 54.60 & 55.70 \\
amh & 28.30 & \underline{36.10} & 33.00 & 34.20 & 31.90 & 35.80 & \textbf{40.40} \\
bel & 49.20 & 49.10 & 47.10 & 49.30 & \underline{49.40} & \textbf{51.10} & 48.60 \\
dan & \textbf{81.40} & 80.00 & 81.00 & \textbf{81.40} & \underline{81.30} & 81.20 & 79.60 \\
deu & 56.80 & \underline{63.10} & 56.40 & 61.30 & 61.70 & \textbf{64.60} & 57.70 \\
eng & 87.70 & 87.80 & 86.60 & 88.10 & 88.30 & \textbf{89.50} & \underline{89.40} \\
fin & \underline{39.30} & 32.10 & 34.70 & 33.40 & \textbf{41.90} & 36.70 & 33.70 \\
fra & 55.70 & 55.20 & 51.50 & 53.90 & \underline{56.20} & \textbf{59.60} & 51.40 \\
grc & 28.00 & 28.20 & 28.30 & \underline{31.70} & 30.60 & \textbf{32.10} & 22.60 \\
heb & \underline{18.43} & 15.01 & 17.22 & 16.31 & \textbf{20.95} & 17.62 & 17.42 \\
hun & 56.70 & 52.70 & 53.90 & \underline{59.80} & \textbf{63.10} & 58.60 & 53.50 \\
hye & 76.10 & 71.90 & 76.10 & \underline{77.30} & \textbf{78.30} & 73.40 & 75.30 \\
ita & 36.30 & 43.90 & 46.10 & \underline{49.00} & 46.30 & \textbf{50.00} & 48.80 \\
jap & 11.20 & 11.20 & 14.60 & 13.30 & \underline{16.80} & \underline{16.80} & \textbf{34.70} \\
rus & 65.20 & 68.90 & 68.70 & 67.40 & \textbf{70.80} & \underline{70.20} & 69.70 \\
san & 21.80 & \underline{27.80} & 24.60 & 25.10 & 20.40 & 25.70 & \textbf{32.30} \\
sme & 31.80 & \textbf{36.80} & 33.90 & \underline{35.70} & 32.90 & \textbf{36.80} & 33.50 \\
spa & 57.40 & 52.20 & 58.70 & \underline{62.60} & 59.50 & \textbf{64.30} & 55.30 \\
tur & 55.70 & 61.70 & \underline{65.60} & 63.80 & \textbf{67.10} & 63.30 & 59.70 \\
Mean & 48.04 & 48.76 & 49.19 & 50.51 & \underline{51.26} & \textbf{51.68} & 50.49 \\
\bottomrule
\end{tabular}
\caption{All language results for the ud-wl-NR dataset.}
\label{tab:wl-r}
\end{table*}

\begin{table*}[t!]
\centering
\begin{tabular}{lllllll}
\toprule
 & eng & hun & ita & rus & spa & Mean \\
\midrule
\texttt{t5-prefix} & 80.10 & 46.60 & 13.90 & 56.50 & 35.30 & 46.48 \\
\texttt{cmlm-prefix} & 88.70 & 56.20 & 34.90 & 63.50 & 47.80 & 58.22 \\
\texttt{t5-suffix} & 88.50 & 64.50 & 44.00 & 60.50 & 49.80 & 61.46 \\
\texttt{cmlm-suffix} & 87.30 & 62.40 & 45.40 & 61.50 & 57.20 & 62.76 \\
\texttt{t5-iid} & 75.90 & 51.10 & 27.40 & 58.20 & 41.20 & 50.76 \\
\texttt{cmlm-iid} & 81.10 & 58.40 & 32.90 & 59.20 & 46.50 & 55.62 \\
\midrule
\texttt{t5-seg-prefix} & 85.90 & 55.20 & 44.90 & 66.30 & 58.60 & 62.18 \\
\texttt{t5-seg-suffix} & \underline{89.60} & 61.60 & \underline{51.90} & 69.20 & 59.30 & 66.32 \\
\texttt{t5-seg-iid} & 89.20 & \underline{66.00} & 48.60 & \underline{71.60} & \underline{61.60} & \textbf{67.40} \\
\texttt{cmlm-seg-prefix} & 85.00 & 56.90 & \textbf{58.20} & 65.40 & 58.40 & 64.78 \\
\texttt{cmlm-seg-suffix} & \textbf{90.00} & 65.60 & 44.20 & \textbf{73.20} & 59.40 & 66.48 \\
cmlm-seg-iid & 89.00 & \textbf{66.10} & 47.80 & 69.60 & \textbf{61.70} & \underline{66.84} \\
\bottomrule
\end{tabular}
\caption{All language results for the segmentation dataset when deleting rather than masking.}
\label{tab:seg-langs-delete}
\end{table*}

\end{document}